\documentclass[runningheads]{llncs}

\usepackage{booktabs}
\usepackage{multirow}
\usepackage{graphicx}
\usepackage{xcolor}
\usepackage{hyperref}
\usepackage{amsmath}
\usepackage{subcaption}
\newcommand{\anon}[1]{\textless\text{#1}\textgreater}
\newcommand{\pred}[2]{\text{#1}(#2)}
\newcommand{\predd}[3]{\text{#1}(#2, #3)}
\newcommand{\lamabs}[2]{\lambda#1.(#2)}

\usepackage{color}
\newcommand{\todo}[1]{}

\begin{document}

\title{Neural Semantic Parsing with\\ Anonymization for Command Understanding in \\General-Purpose Service Robots}

% Abbrieviated
\titlerunning{Command Understanding for General-Purpose Service Robots}

\author{
Nick Walker\orcidID{0000-0001-7711-0003} \and
Yu-Tang Peng\orcidID{0000-0002-4916-5557} \and
Maya Cakmak\orcidID{0000-0001-8457-6610}
}

\authorrunning{N. Walker, Y. Peng and M. Cakmak
}

\institute{Paul G. Allen School of Computer Science \& Engineering \\ 
University of Washington, Seattle WA 98195, USA \\
\email{\{nswalker,pengy25,mcakmak\}@cs.washington.edu}
}

\maketitle          

\begin{abstract}
Service robots are envisioned to undertake a wide range of tasks at the request of users. 
Semantic parsing is one way to convert natural language commands given to these robots into executable representations.
Methods for creating semantic parsers, however, rely either on large amounts of data or on engineered lexical features and parsing rules, which has limited their application in robotics.
To address this challenge, we propose an approach that leverages neural semantic parsing methods in combination with contextual word embeddings to enable the training of a semantic parser with little data and without domain specific parser engineering. 
Key to our approach is the use of an anonymized target representation which is more easily learned by the parser.
In most cases, this simplified representation can trivially be transformed into an executable format, and in others the parse can be completed through further interaction with the user.
We evaluate this approach in the context of the RoboCup@Home \emph{General Purpose Service Robot} task, where we have collected a corpus of paraphrased versions of commands from the standardized command generator.
Our results show that neural semantic parsers can predict the logical form of unseen commands with 89\% accuracy.
We release our data and the details of our models to encourage further development from the RoboCup and service robotics communities.

\end{abstract}

\section{Introduction}

General-purpose service robots (GPSRs) are envisioned as capable helpers that will assist with everything from chores around the home to finding open conference rooms in the office.
These robots are distinguished by their ability to accomplish a wide variety of possible goals by recomposing basic capabilities spanning navigation, manipulation, perception and interaction. 
One especially desirable interface for these robots is natural language, because users are already familiar with using language to ask for assistance.

Command understanding is commonly framed as executable semantic parsing~\cite{chen:aaai11,Matuszek2013}, where the objective is to convert the user command into a logical representation that unambiguously captures the meaning of the command, decoupled from the surface characteristics of the language.
For instance, both ``Bring me a red apple from the kitchen'' and  ``Could you get me a red apple from kitchen please" might be transformed into a $\lambda$-calculus representation like $\pred{bring}{\lamabs{\$1}{\pred{apple}{\$1} \wedge \pred{red}{\$1} \wedge \predd{at}{\$1} {\text{``kitchen''}}}}$.
This representation can then be grounded immediately by finding satisfactory entities from the robot's ontology or later as a part of executing the resulting plan.

Traditionally, creating usable semantic parsers in a low-data domain has required an expert to craft lexical features or parsing rules.
Recent neural semantic parsing approaches have lessened the need for domain specific engineering, but are not easily applied to robotics settings because of the dearth of available annotated data. 
This data-deficit is likely to persist because the contextual nature of commands makes it challenging to collect data without expensive extended interactions, and there is no deployed base of service robots with which interactions can be gathered en masse. 

In this paper, we propose an approach to command understanding in a general-purpose service robot that can make use of recent advances in semantic parsing. 
Key to our approach is the use of a simplified target representation which trades immediate executablitity under certain circumstances for ease of learning.
Further, we leverage pretrained contextual word embeddings to improve the model's generalizability.
We evaluate this approach on a new corpus of command-semantics pairs created by crowd-sourcing paraphrases of the generated language used in the \emph{General Purpose Service Robot} task from RoboCup@Home, demonstrating that learned parsers can predict the correct logical form of an unseen command with 89\% accuracy.

\section{Related Work}

Several works have investigated neural methods for semantic parsing~\cite{misra:emnlp16,kovcisky:emnlp16,lewis:naacl16}.
Our work is based on recent advances in translating language to logical forms using sequence-to-sequence encoder-decoder neural models~\cite{dong2016}. 
Architectures that enforce the constraints of the target representation have been proposed, including recurrent neural network grammars~\cite{dyer-etal-2016-recurrent}, sequence-to-tree methods~\cite{dong2016}, course-to-fine methods~\cite{dong2018}, as well as other applications of constrained decoding tailored for semantic parsing~\cite{krishnamurthy-etal-2017-neural}.
Our work does not enforce decoding constraints, sacrificing potential performance gains for a higher degree of portability across target representations.

Anonymization, also referred to as delexicalization, is frequently used to overcome data sparsity in natural and spoken language systems.
Our work is similar in spirit to the argument identification method used by Dong and Lapata~\cite{dong2016}, but instead of anonymizing entities into unique tokens, we abstract them by type into tokens representing their class.
Perhaps closest to our approach is work on weakly-supervised parser learning aided by abstracted representations~\cite{goldman-etal-2018-weakly}, which similarly proposes leveraging a small lexicon to simplify a visual reasoning domain.
Copy mechanisms~\cite{gu-etal-2016-incorporating}, which enable sequence-to-sequence models to copy portions of their input into their output, are also frequently used to combat data sparsity. 
Recent work has suggested that copying may dominate delexicalization for several semantic parsing tasks~\cite{damonte-etal-2019-practical}.
We do not consider a copy mechanism in this work, as avenues for integrating the robot's knowledge are less available when relying on a learned copying behavior.

The application of contextualized word embeddings to various language tasks is an active area and has seen several early applications to semantic parsing~\cite{einolghozati2019improving}.
To our knowledge, this work is the first to evaluate whether contextual embeddings improve performance for semantic parsing in low-data domains.

Many researchers have investigated methods for giving commands to robots~\cite{tellex:aaai11,Matuszek2013}.
Our work is most closely aligned with research from Thomason et al.~\cite{thomason2019improving} on continually learning a command parsing and grounding system via dialogue.
While we do not consider learning from dialogue interactions, we make the assumption that our understanding system will be used as part of a dialogue agent that can ask for confirmation or corrections.
Further, where Thomason et al. initialized their parser by engineering a CCG lexicon against a set of user commands, we use neural methods that directly learn the correspondence between commands and logical forms.

While many teams have built systems motivated specifically by the RoboCup@Home GPSR task, they have largely adopted techniques that depend on knowledge of the command generation grammar~\cite{matamoros2019}.
Perhaps the most sophisticated system is that of Bastianelli et al., a frame-semantics based spoken language understanding system for service robots which can integrate and utilize visual information~\cite{Bastianelli2014,Bastianelli2016}.
In contrast, our work considers solely the language aspect of command understanding, and takes advantage of the flexible nature of neural translation models to avoid prescribing a particular representation.

\section{Approach}

\begin{figure}
    \centering
      \begin{subfigure}{0.3\textwidth}
  	\includegraphics[width=\columnwidth]{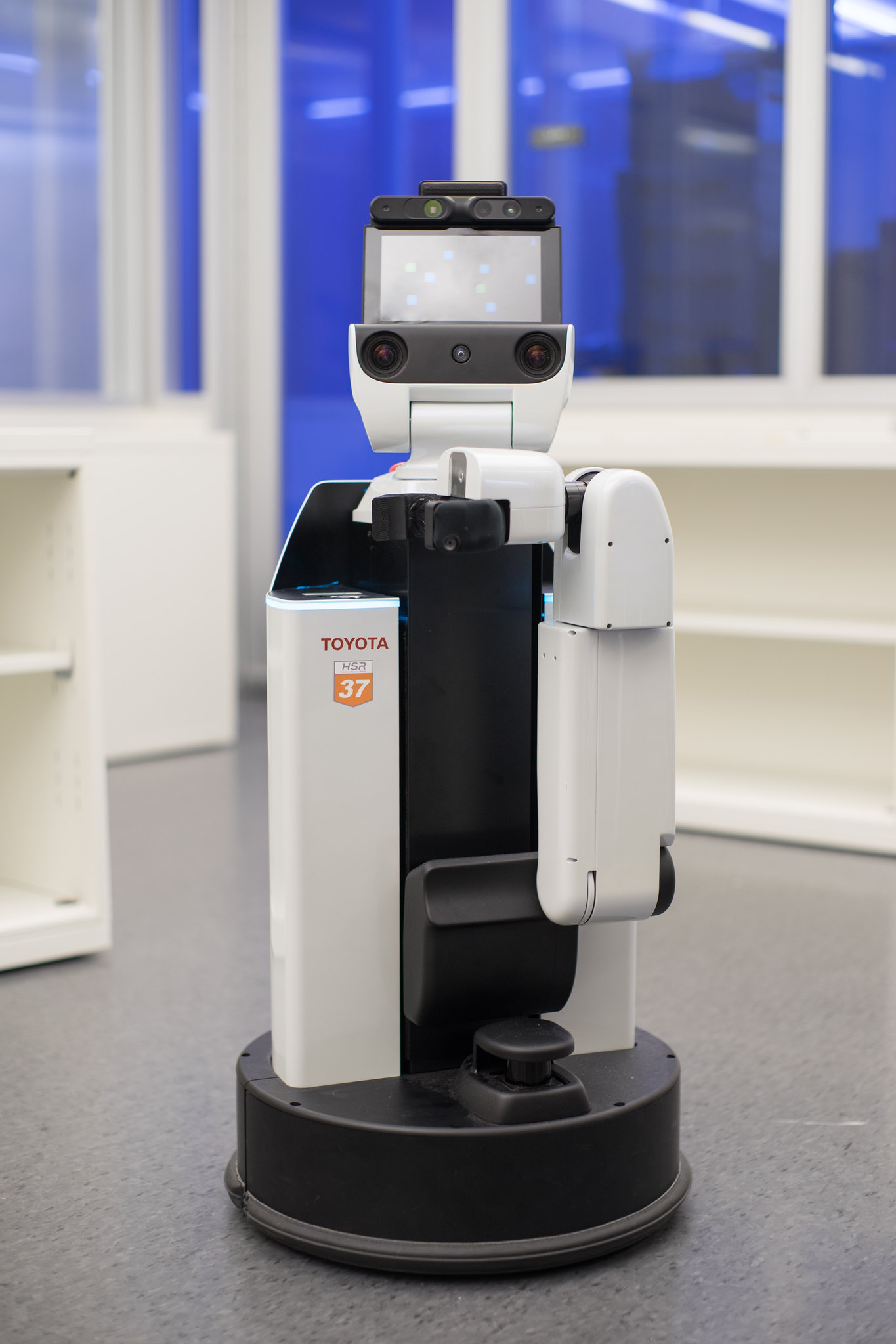}	
  	\caption{}	
    	\label{fig:hsr}
  \end{subfigure}
  \quad
  \begin{subfigure}{0.45\textwidth}
  	\includegraphics[width=\columnwidth]{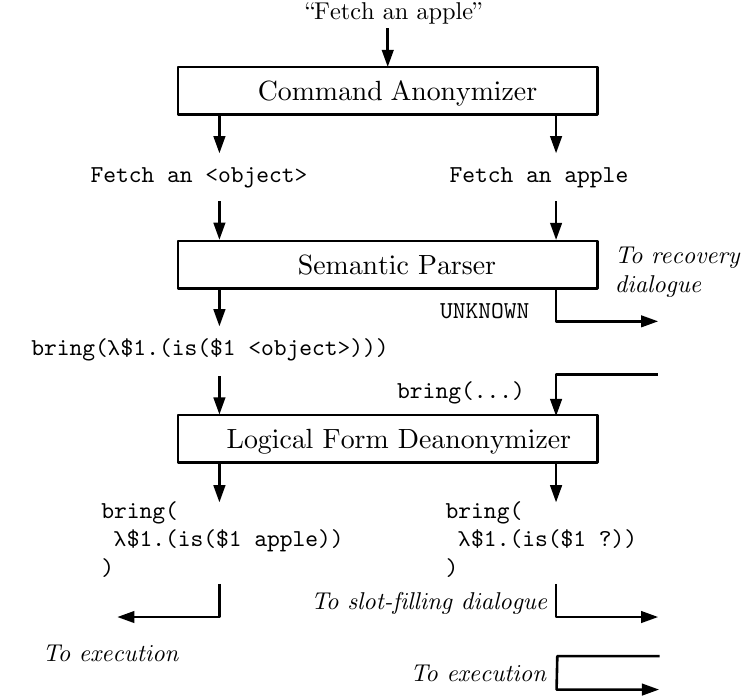}
    \caption{}
    	\label{fig:pipeline}
  \end{subfigure}	
    \caption{ (\subref{fig:hsr}) The Toyota Human Support Robot is used as a standard platform in the RoboCup@Home competition. (\subref{fig:pipeline}) A high level overview of how the approach converts a natural language command to a logical form. The left path traces successful execution of each step. The right path shows how failures can be addressed by gathering additional information via dialogue interaction.}

    \label{fig:overview}
\end{figure}

The thrust of our approach is to leverage the robot's knowledge base to simplify input utterances where possible, then parse them into an abstracted $\lambda$-calculus representation where argument slots are left as tokens representing the class of the argument.
This simplifies the output space of the parser and enables us to use neural semantic parsing, which would otherwise be infeasible due to a lack of in-domain data. 
When important information is lost in this simplification, it can be retrieved easily by asking the user for clarification.
An overview of this framing is shown in Fig.~\ref{fig:overview}.
We augment a standard encoder-decoder model with contextual embeddings to further ameliorate the challenge of data-scarcity.

\subsection{Command Anonymizer}

Though service robots operate in an open world and thus cannot assume complete knowledge of the environment, they are usually equipped with ontologies specifying basic knowledge about objects and locations.
We leverage this to anonymize commands where possible:
Any token in the input command for which we have an name entry in the knowledgebase is replaced with a special token denoting its class. 
For instance, the command ``Fetch an apple from the kitchen" would be anonymized to ``Fetch an \anon{object} from the \anon{location}."

When anonymization is successful, it reduces the complexity of the semantic parsing task.
For commands with arguments that fall within the robot's ontology, the model is no longer required to generalize parses across instantiations with slightly different entities.
Further, anonymization guarantees that newly added entities can be used in the same commands that worked for previous entities.
Because users will frequently refer to previously unknown objects or use unfamiliar language to refer to known objects however, the semantic parser must still be robust to partially- or even completely non-anonymized commands.

\subsection{Semantic Parser}

\begin{figure}
    \centering
    \includegraphics{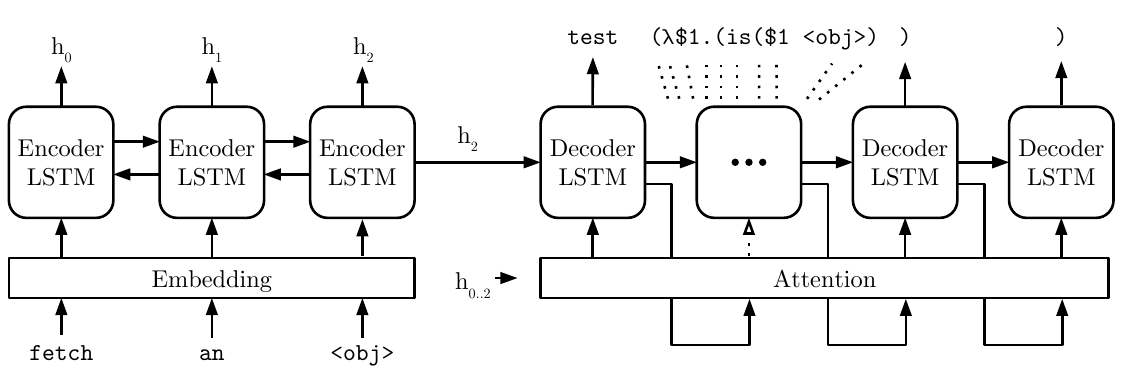}
    \caption{The outline of the seq2seq architecture, shown encoding a command and decoding a logical form. Special start and end tokens are omitted for clarity. Arrows denote representations passing between modules.}
    \label{fig:architecture}
\end{figure}

Following Dong and Lapata~\cite{dong2016}, our parser is a sequence-to-sequence (seq2seq) bidirectional LSTM encoder-decoder model with a bilinear attention mechanism which takes language input and translates it to a logical form.
This architecture is shown in Fig.~\ref{fig:architecture}.
Input tokens are represented with a concatenation of a contextualized word embedding and their 100D GloVe embedding~\cite{pennington2014glove}.
For anonymized commands, the embedding provides a signal for which class of command is being asked for, as the interchangeability of verbs is captured.
For partially anonymized commands, which the model receives when anonymization is unsuccessful, the embeddings may help if the unknown referent phrase embeds near arguments seen during training.

{}

We adopt $\lambda$-calculus for our logical representation, but counter to typical practice where output forms are fully specified, we use an intermediate, anonymized form.
Similar to the anonymized form of a command, arguments to predicates are represented as abstract class tokens that encode the type of the argument but not its identity.
For instance, the command ``Navigate to the kitchen, look for the apple, and give it to Bill" has corresponding anonymized form

\begin{figure}
    \centering
    $\pred{bring}{ \lamabs{\$1}{\predd{is\_a}{\$1}{\anon{object}} \wedge \predd{at}{\$1}{\anon{room}}},\lamabs{ \$1}{\pred{person}{\$1} \wedge \predd{name}{\$1}{\anon{name}} }}$.
\end{figure}

This representation retains the expressive power of $\lambda$-calculus, which can capture compositional aspects of common commands (e.g. ``Bring me the [apple [from the counter [by the refrigerator]]]''), but discards the separate challenge of properly assigning arguments to leaf predicates.
Our observation is that, by applying this abstraction, the space of output logical forms is significantly contracted for a set of representative robot commands, enabling us to train useful models even with relatively little data.

\subsection{Logical Form Deanonymizer}

The output of the semantic parser is an anonymized logical form.
In order for the robot to execute the command however, this form must be deanonymized into a fully specified logical expression.

In cases where the anonymizer had replaced spans in the original input utterance, the deanonymizer has access to the correspondence between the class tokens and the original text that they replaced.
In many cases, this allows the deanonymizer to automatically reconstruct the full logical form.
In some cases, however, there may be multiple possible matchings between class tokens in the anonymized command and the anonymized logical form. 
For instance, for the command ``Move the apple from the kitchen counter to the dining table", there will be two \anon{location} tokens.
Additionally, in cases where anonymization failed, the deanonymizer does not have knowledge of what input spans correspond to the anonymous tokens in the semantic parse. 
Our observation is that these scenarios are easily addressed via dialogue.
Similar to the approach used by Thomason et al.~\cite{thomason2019improving}, we propose that the deanonymizer disambiguate argument assignments via a slot filling dialogue policy.
The same dialogue can be used to add entities to the ontology so they can be properly anonymized in the future.

\section{Data}

To our knowledge, there is no GPSR-style corpus of sufficient scale to evaluate our approach, so we constructed a set of anonymized command-semantics pairs based on the command generator\footnote{The generator is available at \url{https://github.com/kyordhel/GPSRCmdGen}. 
At the time of this work, the 2019 generator had not been finalized.} used in the \textit{General Purpose Service Robot} task from the 2018 international RoboCup@Home domestic service robotics competition~\cite{rulebook_2018}.
In the GPSR task, robots are read a generated command which they must carry out in a mock-apartment arena.
These commands are intended to encompass all of the capabilities that are assessed in other tasks in the competition, including things like finding people and fetching objects.
We supplemented the synthetic data from the generator with paraphrases gathered via crowdsourcing.
In this section, we describe the details of the construction of these datasets.\footnote{The data and splits used for our experiments are available at \url{https://doi.org/10.5281/zenodo.3244800}.}

\begin{table}[]
\renewcommand{\arraystretch}{1.2}
\centering
\caption{Summary of our annotations to the 2018 \emph{General Purpose Service Robot} task command generator, broken down by category. 
``Annotations" refers to the number of logical templates that were annotated atop the grammar rules. 
``Logical forms'' is the number of unique anonymized logical forms produced by expanding the annotations. 
Because categories overlap slightly, we count commands and logical forms as belonging to the first category that they appear in. 
Length is measured in tokens and includes all tokens that would be given or expected from a model.
On average, 58\% of logical form tokens are parentheses, quotation marks, or $\lambda$-calculus variable type markers. }
\label{grammar-annotations}

\begin{tabular}{@{}l@{\hskip 0.25in}rrr@{\hskip 0.25in}r@{}}\\
\toprule
\textbf{Category}             & \textbf{1} & \textbf{2} & \textbf{3} & \textbf{All}  \\ \midrule
Anonymized commands         & 192        & 352        & 667        & 1211          \\
Logical forms               & 17         & 39         & 45         & 101            \\
Annotations                   & 28         & 53         & 44         & 125     \\ \midrule
\textbf{Complexity Measures} \\
Average command length        & 11.9      & 12.2      & 10.4      & 11.3    \\
Average logical form length   & 28.4      & 27.0      & 22.6      & 25.1         \\
Commands to forms ratio & 11.3       & 9.0       & 14.8       & 12.0          \\\bottomrule
\end{tabular}

\end{table}

\subsection{Generated}

The generator is specified as a probabilistic context-free grammar with equal weighting on all production rules, split into three categories. 
Each category introduces both more complicated commands as well as more complicated desired goals.
Example commands are provided for each category in Table~\ref{grammar-examples} and summary information about the grammar is shown in Table~\ref{grammar-annotations}. 

The generator is distributed with an ontology describing objects, object categories, person names, gestures, question-answer pairs, common sayings, and household locations.
Because commands have multiple highly-branching non-terminals representing entities from the ontology, the number of full expansions grows exponentially in the size of the ontology.
However, we observe that the number of distinct anonymized commands is actually quite low, and the number of distinct logical forms is lower still.

We defined a logical domain consisting of 27 predicates (7 actions, 20 descriptive) and used them to create 125 annotations that provide logical forms for all 1211 distinct anonymized commands. 
These predicates cover concepts such as names, basic prepositions (e.g. \texttt{at}, \texttt{left\_of}), and entity types (e.g. \texttt{person}, \texttt{object}).
%Because part of the grammar explicitly incomplete commands, we include an \texttt{UNKNOWN} token to represent the value for a part of the parse that cannot adequately resolved from the command alone.

There is no general correspondence between the structure of the generation grammar and the structure of the resulting logical representation; however expansions near the leaves frequently take forms that can be neatly tagged with a semantic template. 
Thus we annotate partial expansions from the generator with an accompanying semantic template which contains some of the same non-terminals, creating a shallow synchronous context free grammar that produces both commands and their logical representations when expanded.

For instance, a partial expansion that produces ``bring" commands contains a nonterminal which produces synonymous verbs and another non-terminal which produces different objects.
Its annotation incorporates the object non-terminal but discards the verb non-terminal as its expansions do not affect the semantics of the command:

\begin{equation*}
     \text{\$vbbring}\ \text{me\ the}\ \text{\$object} = \pred{bring}{\lamabs{\$1}{\predd{is\_a}{\$1}{\$object}}}
\end{equation*}

Continuing to expand both the command and this semantic template will result in a fully specified command-semantics pair.
To produce pairs of anonymized commands and anonymized logical forms, we simply modify productions associated with ontology entities to produce our class tokens.

\subsection{Paraphrases}

Although the GPSR command generator is designed with naturalism in mind, it is nonetheless artificial.
In order to understand how our approach will fair given more realistic input, we applied a similar methodology as that of Wang, Berant and Liang to crowdsource paraphrases of our generated dataset~\cite{wang-etal-2015-building}.
Crowd workers were provided fully-expanded generated commands and prompted to provide a paraphrased version---new text that captures all of the same information but uses different words or phrasing.
We nudged workers to provide substantial paraphrases by presenting a warning UI if their input was below a threshold of Levenshtein (character) and Jaccard (word) distances from the original command.
A similar warning was presented if the same metrics indicated that the paraphrase contained almost no overlap, as this almost always indicated that important information was discarded.
To help filter out low-quality responses, we required crowd-workers to write their own commands based on their impression of what the robot could do.
Spam responses to these questions consistently indicated spam responses to the paraphrasing task.

We used Amazon Mechanical Turk to collect 1836 paraphrases from 95 workers, ensuring there were at least 10 paraphrases per logical form.
The mean Levenshtein and Jaccard distances between a paraphrase and its original command are 28.0 and 0.59 respectively.
Sample paraphrases are shown in Table~\ref{grammar-examples}.

\begin{table}[]
\renewcommand{\arraystretch}{1.09}
\centering
\caption{Examples of fully-expanded commands, their anonymized forms, crowd-sourced paraphrases, and corresponding logical forms from each category of the GPSR task.}
\label{grammar-examples}
\begin{tabular}{@{}c l@{}}\\
    \toprule \textbf{Category} & \textbf{Example}\\ \midrule
   \multirow{4}{*}{\textbf{1}}     & tell me how many coke there are on the freezer\\ 
   & tell me how many \anon{object} there are on the \anon{location}\\
    & ``how many cokes are left in the freezer"\\
    &  $\pred{say}{ \pred{count}{ \lamabs{\$1}{ \predd{is\_a}{\$1}{\anon{object}}} \wedge \predd{at}{\$1}{\anon{location}}}}$\\
    \midrule
    \multirow{4}{*}{\textbf{2}} & tell me what's the largest object from the bar \\
    & tell me what's the largest object from the \anon{location}\\
    & ``Which is the largest object on top of the bar" \\
    & $\pred{say}{\lamabs{\$1}{\pred{largest}{\$1} \wedge \predd{at}{\$1}{\anon{location}}}}$\\
    \midrule
    \multirow{4}{*}{\textbf{3}} & Could you give me the object on top of the glass from the coffee table \\
    & Could you give me the object on top of the \anon{object} from the \anon{location} \\
    & ``can you bring be the thing from the coffee table thats on top of the glass" \\
    & $\predd{bring}{\lamabs{\$1}{\lamabs{\$2}{\predd{is\_a}{\$2}{\anon{object}} \wedge \predd{on\_top\_of}{\$1}{\$2}} }}{\anon{location}}$ \\
    \bottomrule

\end{tabular}
\end{table}

\section{Experiments} \label{experiments}

We evaluate whether it is feasible to learn usable semantic parsers under our approach given the relatively small amount of data available.
Our metric is exact-match accuracy; the percentage of logical forms that a model predicts exactly correctly on held-out test data.

Though we are primarily interested in how well models perform on the paraphrased data---as this best represents real language that a robot may encounter--- we take advantage of the generated-paraphrased corpus' parallel nature to investigate a range of interesting configurations:

\begin{enumerate}
    \item \emph{Train generated, test generated} lets us see whether a learned model can approximate the performance of a grammar-based chart-parser, even without having access to the full grammar.
    \item \emph{Train generated, test paraphrased} exposes the extent to which the generator captures aspects of natural commands.  
    \item \emph{Train paraphrased, test paraphrased} shows a model's capacity to generalize across real commands.
    \item \emph{Train generated and paraphrased, test paraphrased} can reveal whether there are any benefits to augmenting the paraphrased data with synthetic language.
\end{enumerate}

To better approximate the conditions of a real service robot deployment, we do not use prior knowledge of entities in our experiments.
Generated data consists purely of pairs of anonymized commands and anonymized logical forms.
Paraphrased commands are not anonymized before being processed by the model.

We split both the generated data and the paraphrased data 70\%/10\%/20\% into training, validation and test sets.
Splits are such that no command appears in more than one set.
As noted by Finegan et al., testing only generalization to unseen commands can reward models that learn to simply classify commands and produce memorized logical forms~\cite{finegan2018}.
To evaluate whether models are able to produce correct, unseen logical forms, we use an additional logical split.
This split forces the pools of logical forms in each part of the dataset to be disjoint.
We ensure that the logical split roughly matches the proportions of the command split, and that the split is synchronized across generated and paraphrased data so that they can be combined without causing data leakage.

\subsection{Training Regime} \label{training}

We use the Adam optimizer to minimize the cross entropy of the predicted output with the ground truth labels.
Training is performed for 150 epochs with early stopping (patience=10) based on accuracy evaluation on the validation set.
We use an encoder dropout probability of .1.
Test- and validation-time decoding is performed using beam search with a width of 5.
Our experiments are built on top of AllenNLP~\cite{Gardner2017AllenNLP} which uses the PyTorch deep learning framework.\footnote{The details of our implementation, including the full parameterization of our experiments, is available at \url{https://doi.org/10.5281/zenodo.3246755}.}

We provide results using each of ELMo~\cite{peters2018}, OpenAI GPT1~\cite{radford2018}, $\text{BERT}_{\textit{base}}$, $\text{BERT}_{\textit{large}}$~\cite{devlin2018bert}, with a comparison against a model that forgoes a contextual word embedding and simply uses GloVe. 
We leave the contextual word embedding frozen during training and instead allow tuning of the GloVe weights to avoid catastrophic forgetting.

\subsection{Baseline Models}

We compare our models against two simple baselines.
\textsc{Grammar-Oracle} chart parses test samples using the generation grammar and looks up the corresponding annotation to return a logical form.
Because it always has full access to the grammar, its predictions depend only on whether the test data are within the grammar.
The K-nearest neighbors (\textsc{KNN}) model predicts the label of a test command by searching for its nearest neighbor amongst the training and validation data.
We found empirically that using Jaccard distance and $K=1$ worked well. 
This model is naturally incapable of predicting labels it has never seen before, so it always scores 0\% when evaluated on a logical split. 

\section{Results}

The results of our experiments are shown in Table~\ref{results}.
The columns of the table and our discussion are ordered to match the sequence of the descriptions given in Section~\ref{experiments}.

\begin{enumerate}
\item When trained and evaluated on synthetic data, neural semantic parsing models easily fit to the surface characteristics of the data, achieving accuracy levels of 98.8\% on unseen commands, despite not having access to the underlying grammar.

\item Models trained with the generated data achieve at most 27.6\% accuracy on unseen paraphrased commands.
All models are bested by the \textsc{KNN} baseline.
As the generated training data doesn't contain any entities, it is unsurprising that test performance on the completely non-anonymized paraphrasing data is poor.
The results indicate that around a quarter of the paraphrased commands can be resolved purely by looking for structural patterns learned from the generated language.

\item The best model trained on paraphrased data alone is able to achieve 78.5\% accuracy on unseen commands, indicating that neural semantic parsers are reasonably capable of handling a realistic GPSR task.
Comparing the best BERT-based model against the model with no contextual word embedding indicates that these large pretrained models can provide around an 8\% performance improvement for this task.

\item Training with both the generated and paraphrased data leads to a consistent and large performance boost across all models, yielding the best performing model as evaluated on the paraphrasing test set.
This result is possibly a reflection of headroom for improvement if more real data were available to the model.
\end{enumerate}

Mirroring previously reported results on text-to-SQL tasks~\cite{finegan2018}, the generalization achieved across unseen commands does not extend to unseen logical forms.

\begin{table}[]
\renewcommand{\arraystretch}{1.2}
\newcommand{\mc}[3]{\multicolumn{#1}{#2}{#3}}
\newcommand{\bes}[1]{\textbf{#1}}
\centering
\caption{Accuracy of models and ablations on different datasets. ``C" indicates results from data split on commands while ``$\lambda$'' indicates results from data split on logical forms.}
\label{results}
\begin{tabular}{@{}l rr c@{\hspace{.4cm}} rr c@{\hspace{.4cm}} 
rr c@{\hspace{.4cm}}  rr@{}}\\
\toprule
Train                  
& \mc{5}{c}{Gen.}   && \mc{2}{c}{Para.} && \mc{2}{c}{G. + P.} \\
\cmidrule(lr){2-6} \cmidrule(lr){8-9} \cmidrule(lr){11-12}
Test                   
& \mc{2}{c}{Gen.}     &&  \mc{2}{c}{Para.}       &&  \mc{2}{c}{Para.}  && \mc{2}{c}{Para.}\\ 
\cmidrule(lr){2-3} \cmidrule(lr){5-6}  \cmidrule(lr){8-9} \cmidrule(lr){11-12}
Split                 
& \mc{1}{c}{C} & \mc{1}{c}{$\lambda$} &&  \mc{1}{c}{C} & \mc{1}{c}{$\lambda$}  && \mc{1}{c}{C} & \mc{1}{c}{$\lambda$} && \mc{1}{c}{C} & \mc{1}{c}{$\lambda$}\\
\midrule
\textsc{Grammar-oracle}        
& 100.0  & 100.0    && 1.1 & 0.9       && 1.1 & 0.9       && 1.1  & 0.9 \\

\textsc{KNN}     
& 63.0  & 0.0       && \bes{42.0} & 0.0 && 42.8   & 0.0   && 49.8 & 0.0 \\
\textsc{seq2seq}      
& 95.9   & 0.0      && 13.0 & 0.0       && 64.4 & 0.0       && 79.6 & 0.0 \\
\hspace{1mm}+ GloVe       
& \bes{98.8}   & 26.3       && 12.4 & 0.0      && 70.2 & 0.0       && 85.3 & 6.3                                 \\    
\hspace{1mm}+ GloVe;ELMo    
& \bes{98.8}   & 36.2    && 21.3 & 0.0        && 77.3 & 22.1     && 85.4 & 34.4       \\
\hspace{1mm}+ GloVe;OpenAI  
& 97.9  & 24.6     && 27.6 & 1.8             && 78.2 & 26.3    && 89.0 & \bes{37.9}\\
\hspace{1mm}+ GloVe;$\text{BERT}_{\textit{base}}$
& 96.3  & \bes{56.9}   && 12.2 & 3.3            && 75.4 & \bes{31.3}       && 87.6 & 37.6\\
\hspace{1mm}+ GloVe;$\text{BERT}_{\textit{large}}$
& 97.9  & 54.7      && 27.1 & \bes{9.6}     && \bes{78.5} & 30.4      && \bes{89.4} & 37.6\\
\bottomrule        
\end{tabular}

\end{table}

\subsection{Error Analysis}
We manually inspected the predictions of the best paraphrase parsing model, seq2seq with GloVe and $\text{BERT}_{\textit{large}}$, to better understand the model's generalization and common errors.

\subsubsection{Unexpected defaults}
\begin{table}[]
    \centering
    \caption{Multiple inputs mapping to the same, incorrect logical form}
    \begin{tabular}{@{}l  p{6cm}@{}}\\
     $x_1$ & ``Bring an umbrella" \\
     \midrule
     $x_2$ & ``Navigate to the hallway" \\
     \midrule
     $x_3$ & ``Do this then that" \\
    \midrule
     $y_1 = y_2 = y_3$ & ( go `` \anon{room} " ) \\
    \end{tabular}
    \label{tab:default_behavior}
\end{table}
As shown in Table~\ref{tab:default_behavior}, we observed that disparate commands can produce the same prediction.
One explanation is that no probable solutions are found during decoding, so output tends to a default that is helpful for fitting the training data.
Sampling additional commands that yield \texttt{UNKNOWN} might address this.
Alternatively, a confidence threshold could be used to discard low scoring decodings.

\subsubsection{Overly sensitive}
\begin{table}[]
    \centering
    \caption{Predictions can change based on a single word}
    \begin{tabular}{@{}l  p{10cm}@{}}\\
     $x_4$ &  ``Bring an umbrella to me'' \\
     $y_4$ & ( bring ( $\lambda$ \$1 e ( is\_a \$1 ``\anon{object}" ) ) ) \\
     \midrule
     $x_5$ & ``Bring an umbrella to Bob'' \\
     $y_5$ & ( go ``\anon{room}" ) \\
     \midrule
     $x_6$ & ``Bring an umbrella to him'' \\
     $y_6$ & ( guide ( $\lambda$ \$1 e ( person \$1 ) ( name \$1 ``\anon{name}" ) ) ``\anon{location}" ) \\
    \end{tabular}
    \label{tab:word_difference}
\end{table}

As illustrated in Table~\ref{tab:word_difference}, inputs that differ by a single word can be parsed to drastically different logical forms.
Thus we suspect that the model put undue emphasis on the different arguments when predicting the first predicate.

\section{Discussion}

We have shown that the task of understanding commands in a general-purpose service robot domain can be successfully addressed using neural semantic parsing methods.
Key to this success is the use of contextual word embeddings and a abstracted target representation which simplifies the learning task.
Though anonymization may require a clarification dialogue for complex commands, such a dialogue would, in practice, occur regardless simply to confirm the command.

The dataset produced for this work is a strong common basis for command-taking in general-purpose service robots.
Though the task and the generator are set in the home, the types of goals that arise from the commands apply to a wide variety domains.
Thus, we expect that this data can be used to accelerate future efforts to bootstrap command understanding systems for service robots.
We hope that the ease with which it is possible to train capable systems under this framework will motivate members of the RoboCup@Home and service robotics communities to expand their expectations of robot language understanding systems.

\subsubsection{Acknowledgements}

We thank Yuqian Jiang, Jesse Thomason, and the anonymous reviewers for their helpful feedback.
This work was supported by HONDA award ``Curious Minded Machines'' and the National Science Foundation award IIS-1552427 ``CAREER: End-User Programming of General-Purpose Robots."

\bibliographystyle{splncs04}
\bibliography{references}

\end{document}